%% file: main.tex
\pdfoutput=1

\documentclass[11pt]{article}

\usepackage[preprint]{acl}

\usepackage{times}
\usepackage{latexsym}

\usepackage[T1]{fontenc}

\usepackage[utf8]{inputenc}

\usepackage{microtype}

\usepackage{inconsolata}

\usepackage{graphicx}

\usepackage{amsfonts}
\usepackage{mathtools}
\DeclareMathOperator*{\argmax}{arg\,max}
\usepackage[linesnumbered,ruled,vlined]{algorithm2e}

\usepackage{algpseudocode}
\usepackage{algorithmicx}
\usepackage{algcompatible}
\usepackage{xcolor}
\usepackage{multirow}
\usepackage{colortbl}
\usepackage{adjustbox}
\usepackage{tabularx, booktabs}
\usepackage{caption}
\usepackage{subcaption}
\usepackage{tcolorbox}
\usepackage{enumitem}
\usepackage{hyperref}

\usepackage{amsmath}
\usepackage{xspace}
\usepackage{makecell}

\title{Semantic Exploration with Adaptive Gating \\ for Efficient Problem Solving with Language Models}

\author{
 \textbf{Sungjae Lee\thanks{Equal Contribution}\textsuperscript{1}},
 \textbf{Hyejin Park\footnotemark[1]\textsuperscript{2}},
 \textbf{Jaechang Kim\textsuperscript{2}},
 \textbf{Jungseul Ok\thanks{Corresponding Author}\textsuperscript{1,2}},
\\
\\
 \textsuperscript{1}Department of Computer Science and Engineering, POSTECH, South Korea \\
 \textsuperscript{2}Graduate School of Artificial Intelligence, POSTECH, South Korea
\\
{\texttt{\{sungjaelee25,parkebbi2,jaechang,jungseul\}@postech.ac.kr}}
 }

\begin{document}
\maketitle
\begin{abstract}

Recent advancements in large language models (LLMs) have shown remarkable potential in various complex tasks requiring multi-step reasoning methods like tree search to explore diverse reasoning paths. However, existing methods often suffer from computational inefficiency and redundancy. First, they overlook the diversity of task difficulties, leading to unnecessarily extensive searches even for easy tasks. Second, they neglect the semantics of reasoning paths, resulting in redundant exploration of semantically identical paths. To address these limitations, we propose Semantic Exploration with Adaptive Gating (SEAG), a computationally efficient method. SEAG employs an adaptive gating mechanism that dynamically decides whether to conduct a tree search, based on the confidence level of answers from a preceding simple reasoning method. Furthermore, its tree-based exploration consolidates semantically identical reasoning steps, reducing redundant explorations while maintaining or even improving accuracy.  
Our extensive experiments demonstrate that SEAG significantly improves accuracy by 4.3\% on average while requiring only 31\% of computational costs compared to existing tree search-based methods on complex reasoning benchmarks including GSM8K and ARC with diverse language models such as Llama2, Llama3, and Mistral. 
Our code is available at \href{https://github.com/ml-postech/SEAG-semantic-exploration-with-adaptive-gating}{https://github.com/ml-postech/SEAG-semantic-exploration-with-adaptive-gating}.

\end{abstract}

\input{contents/introduction}

\input{contents/related_work}
\input{contents/preliminary}

\input{contents/method}
\input{contents/experiment}

\input{contents/conclusion}

\input{contents/limitation}

\input{contents/ack}



\bibliography{custom}

\clearpage

\appendix

\input{contents/appendix}

\end{document}

%% file: contents/introduction.tex
\section{Introduction}

\begin{figure*}[ht!]
     \centering
    \includegraphics[width=\textwidth]{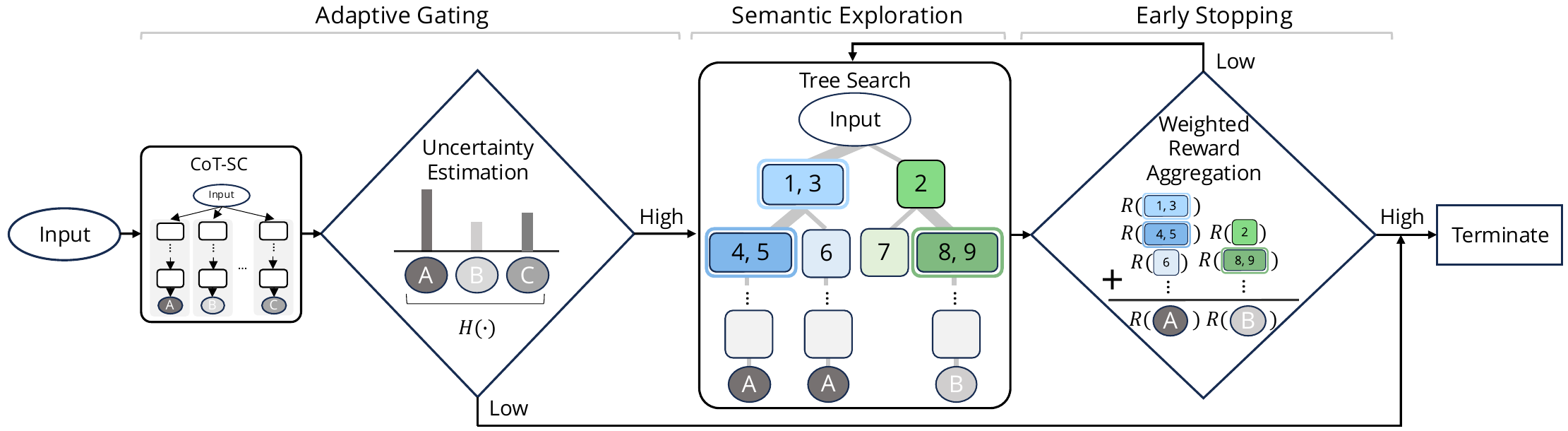}
    \caption{An overview of SEAG. The framework consists of three main components: adaptive gating, semantic exploration and early stopping, detailed in Section~\ref{sec:method}.
    Given an input, \textbf{adaptive gating} determines whether to expand the search space based on the uncertainty $H(\cdot)$ in equation~\eqref{eq:ent} of the generated answers obtained through multiple single-path reasoning.
    If the uncertainty is high, \textbf{semantic exploration} employs tree search to explore multiple reasoning paths, where actions are grouped into semantic clusters (further detailed in Figure~\ref{fig:SE}). Lastly, \textbf{early stopping} terminates the search when 
    the highest aggregated reward surpasses a predefined threshold.}\label{fig:pipeline}
    \vspace{-1em}
\end{figure*}

Recent advances in Large Language Models (LLMs)~\cite{brown2020language, chowdhery2023palm, team2023gemini, touvron2023llama2, achiam2023gpt} have demonstrated remarkable potentials in a wide range of complex reasoning tasks including mathematical problem-solving~\cite{lewkowycz2022solving, wu2022autoformalization, mishra2022lila}, knowledge 
application~\cite{zhang2018variational, yavuz2022modeling}, and commonsense reasoning~\cite{patel2021nlp, sanh2022multitask, madaan2022language}.
LLMs have made notable strides in complex reasoning tasks~\cite{nye2021show, gao2023pal}, yet they face significant challenges in multi-step reasoning. 

Building on a single-path prompting, such as Chain-of-Thought (CoT) and its variants~\cite{wei2022chain, wang2023self, kojima2022large}, recent research has investigated tree search-based methods~\cite{yao2023tree, long2023large,hao2023reasoning} for complex tasks that demand exploration in reasoning.
However, despite their effectiveness, tree search-based methods often face significant computational inefficiencies caused by two primary factors.
First, tree search-based methods are often used for tasks that do not require such complex exploration. 
However, a key challenge lies in determining when to use the single-path method versus tree-based exploration.
Second, the search process repeatedly expands and explores semantically redundant reasoning paths. Addressing the redundancy can be challenging, as it involves identifying when different natural language expressions convey the same meaning~\cite{kuhn2023semantic, farquhar2024detecting}, especially in open-ended reasoning.

To address these limitations, we propose Semantic Exploration with Adaptive Gating (SEAG), a novel framework to significantly enhance computational efficiency in complex reasoning tasks. As illustrated in Figure~\ref{fig:pipeline}, 
SEAG operates in three key phases: adaptive gating (AG), semantic exploration (SE), and early stopping. The first phase, AG (Section~\ref{sec:method:ag}) evaluates the confidence of answers produced by simpler reasoning methods, such as CoT-SC. Based on this confidence, AG determines whether further tree-based exploration is needed, ensuring efficient use of computational resources.
Building on this, the second phase, SE (Section~\ref{sec:method:se}) groups similar reasoning steps using semantic clustering~\cite{kuhn2023semantic, farquhar2024detecting}, avoiding repeated exploration of semantically equivalent paths. 
This approach effectively handles instances where different expressions convey the same meaning, as illustrated in Figure~\ref{fig:SE}.
Additionally, SE prioritizes exploration of semantic clusters containing more instances to derive solutions with higher potentials.
In the final phase, early stopping (Section~\ref{sec:method:es}) terminates the tree search once a highly confident solution is found, avoiding unnecessary computations. SEAG aggregates results by assigning higher importance to reasoning paths that are more semantically relevant or frequently supported. 



Our main contributions are as follows:
\begin{itemize}[leftmargin=13pt, topsep=0pt, itemsep=0pt, parsep=3pt]   \item We propose semantic exploration, leveraging semantic clustering to reduce redundant computations and prioritize exploration of semantically informative reasoning paths  (Section~\ref{sec:method:se}).
    \item 
    We introduce a unified approach combining adaptive gating (Section~\ref{sec:method:ag}) and early stopping (Section~\ref{sec:method:es}) to adaptively determine when to initiate and terminate tree-based exploration, efficiently utilizing computational resources based on confidence measures specific to each process.
    \item Through extensive experiments on multi-step reasoning benchmarks, SEAG consistently demonstrates superior performance in reasoning accuracy while achieving computational efficiency comparable to or greater than baselines (Section~\ref{sec:exp:result}), as shown in Figure~\ref{fig:scatter:llama3}.
\end{itemize}

%% file: contents/related_work.tex
\section{Related Work}

\begin{figure*}[htb!]
     \centering
     \hfill
     \begin{subfigure}[b]{0.45\textwidth}
         \centering
         \includegraphics[width=\textwidth]{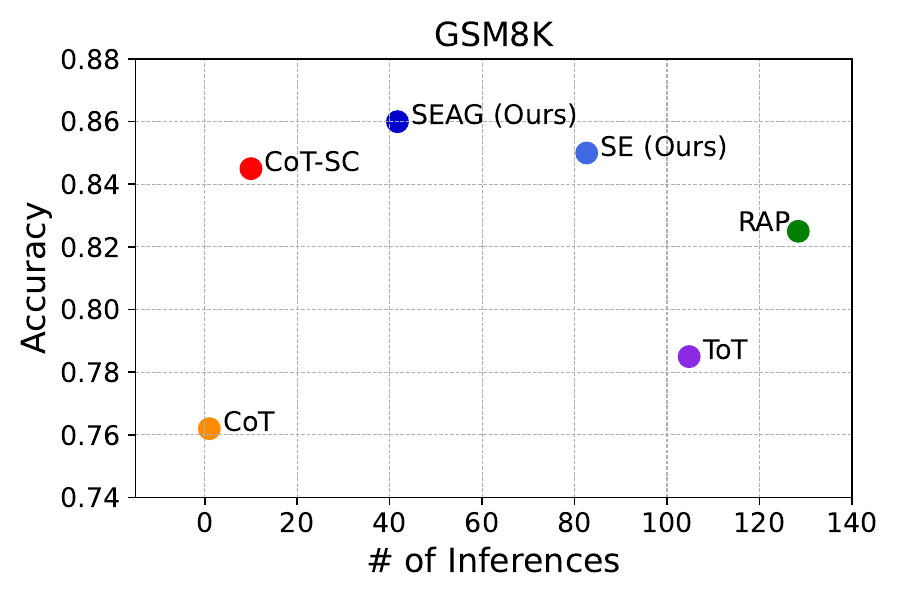}
     \end{subfigure}
     \hfill
     \begin{subfigure}[b]{0.45\textwidth}
         \centering
         \includegraphics[width=\textwidth]{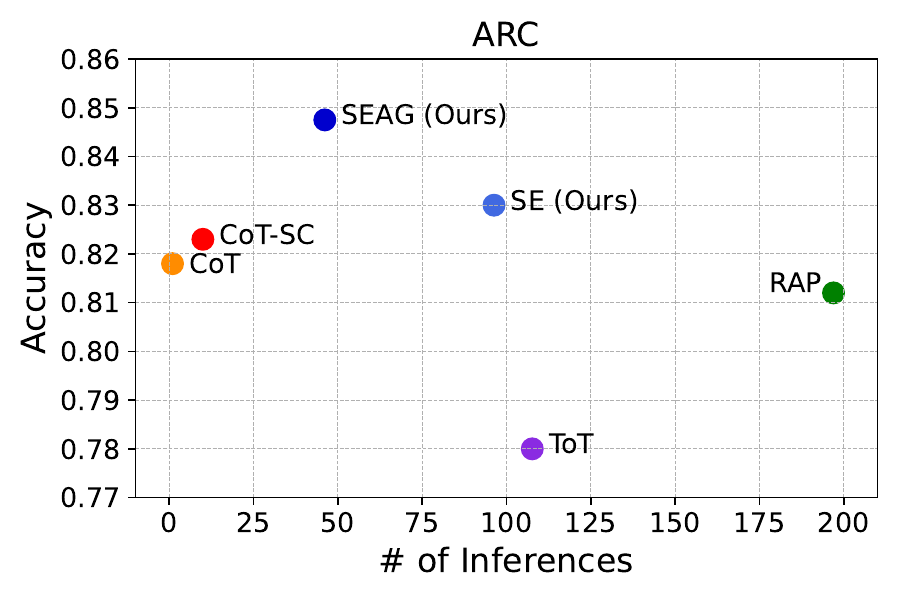}
     \end{subfigure}
     \vspace{-4mm}
     \hfill
        \caption{Scatter plots of accuracy and the number of LLM inferences for baselines and our methods, SE and SEAG, with GSM8K (left) and ARC (right) using Llama3-8B-Instruct model. SE achieves superior accuracy while reducing inference cost compared to existing tree search-based methods. SEAG further improves performance, achieving higher accuracy with efficiency comparable to other reasoning methods.}
        \label{fig:scatter:llama3}
        \vspace{-1em}
\end{figure*}

\paragraph{Reasoning with LLMs}
Eliciting the reasoning capabilities of LLMs has been a key focus of recent research, leading to various reasoning methods to improve their performance on complex, multi-step tasks. One notable approach is Chain-of-Thought (CoT) prompting~\cite{wei2022chain}, which encourages LLMs to generate intermediate reasoning steps. Self-consistency of CoT (CoT-SC)~\cite{wang2023self} extends CoT by sampling multiple reasoning paths and selecting the most frequently answers, instead of relying on a single greedy decoding pass.

Tree search-based approaches, such as Tree-of-Thoughts (ToT)~\cite{yao2023tree, long2023large} and Reasoning-via-Planning (RAP)~\cite{hao2023reasoning}, enhance reasoning by structuring the exploration of reasoning paths on trees using search algorithms such as breadth/depth-first search (BFS/DFS) and Monte Carlo Tree Search (MCTS). 
In addition, recent studies on MCTS-based reasoning have extended the settings involving additional training costs \cite{wan2024alphazero, zhang2024rest} or external feedback \cite{zhou2024language}. 
However, these tree search-based methods incur considerably higher computation costs compared to simpler methods such as CoT-SC.


\paragraph{Linguistic semantics} 
Measuring semantic relation is essential for reducing semantic redundancy in LLM reasoning. Traditional approaches have relied on lexical feature-based~\cite{fernando2008semantic, socher2011dynamic} or embedding-based similarity metrics~\cite{yu2014deep, issa2018abstract}. However, these methods often lack robustness in capturing deeper liguistic semantics. With the rise of Transformer-based models and LLMs, semantic evaluation has achieved significant improvements~\cite{he2020realformer, tay2021charformer}. Recently, \citet{kuhn2023semantic} has introduced a textual entailment-based method for assessing semantic equivalence, which informs our approach to semantic clustering.

Similar to our work, \citet{jang2021monte} has utilized semantic similarity within a MCTS framework for text-based games. However, their method relied on a predefined set consisting of valid actions provided by the game environment. In contrast, we address open domain problems, where actions, such as task-specific sub-questions, are dynamically generated by prompting LLMs.

\paragraph{Uncertainty estimation} 
Uncertainty estimation of LLMs has emerged as a critical area of study for evaluating and improving their reliability. Existing uncertainty estimation methods often rely on the consistency of sampling results~\cite{wang2023uncertaintyawareparameterefficientselftrainingsemisupervised, cole2023selectively} or responses to semantically equivalent questions~\cite{zhang2024sac3reliablehallucinationdetection}. These approaches share an assumption that uncertainty can be quantified through entropy of prediction, and the assumption is commonly used in traditional machine learning~\cite{malinin2018predictiveuncertaintyestimationprior}.
We integrate uncertainty estimation directly into the reasoning pipeline, whereas other methods typically treat it as a separate step.

%% file: contents/preliminary.tex
\section{Preliminaries}

In this section, we first define our problem setting as Markov Decision Process (MDP) reasoning in Section~\ref{sec:mdp}. Section~\ref{sec:mcts} explains Monte Carlo Tree Search (MCTS) for multi-step reasoning.

\subsection{Markov Decision Process Reasoning}\label{sec:mdp}
Let $p_{\theta}$ denote a pre-trained language model (LM) parameterized by $\theta$, and an input sequence $x = (x_1, \ldots, x_{l_x})$, where $l_x$ represents the token lengths of the input.
The model's probability of generating $x$ is expressed by $p_{\theta}(x)=\prod\nolimits_{i=1}^{l_x}(x_i|x_{<i})$, where $x_{<i}$ represents the sequence of tokens preceding $x_i$.
An output sequence $y = (y_1, \ldots, y_{l_y})$ is generated by auto-regressively, where $l_y$ denotes the token lengths of the output. The previously generated tokens are used to predict the next token as $p_\theta(y|x) = \prod_{i=1}^{l_y} p_\theta(y_i | x, y_{<i})$.

Rather than directly mapping the input $x$ to the output $y$, several reasoning methods have been developed to enhance reasoning by breaking down complex tasks into step-by-step thoughts, with each generated token $y_i$ representing an intermediate reasoning step.
Detailed explanations of these methods are provided in Appendix~\ref{append:reasoning}.
Following the approach in RAP~\cite{hao2023reasoning}, we define our problem as an MDP to effectively model the reasoning task.
We leverages LMs in two complementary roles: (i) a reasoning agent and (ii) a world model.
This approach frames the reasoning process as an MDP, enabling iterative reasoning through planning and simulation.
At each time step $t$, the state $s_t$ represents the current context, including both the input sequence and the reasoning history.
In the first role, the LM acts as the reasoning agent, generating actions $a_t \sim p_\theta(a|s_t, m)$ based on the current state $s_t$ and a task-specific instruction $m$. 
Subsequently, the LM functions as a world model, predicting the next state $s_{t+1}$ based on the current state and the chosen action, i.e., $p_\theta(s_{t+1}|s_t, a_t, m')$, where $m'$ is an additional prompt guiding the transition process.
Each reasoning step is evaluated using a reward function $r_t = r(s_t, a_t) \in \mathbb{R}$, based on the LM's self-evaluation of the generated action and the log probability of the action with the history of states. Further details of the reward design are explained in Appendix~\ref{append:reward}.

\subsection{Monte Carlo Tree Search (MCTS)}\label{sec:mcts}
To enhance strategic exploration, we incorporate planning with MCTS~\cite{coulom2006efficient}, which is particularly effective for decision-making in high-dimensional spaces.
MCTS  builds a search tree over $k$ iterations to explore decision space through four main operators: selection, expansion, simulation, and back-propagation.
At each node, which represents a state $s$, the standard MCTS uses UCT~\cite{kocsis2006bandit, auer2002finite} algorithm to select the optimal action $a^*$ based on $Q$-value as follows:
\begin{align}
    a^*=\argmax_{a \in A(s)} \left(Q(s,a)+w \sqrt{\frac{\ln{N(s)}}{N(s,a)}} \right) \;,
\end{align}\label{eq:uct}
where $N(\cdot)$ denotes the total number of visits in previous iterations, $A(s)$ is the set of possible actions at state $s$, and $w$ is the constant of balancing exploration and exploitation.

PUCT~\cite{silver2016mastering, silver2017mastering} 
provides a viable alternative to UCT by incorporating $\pi(a|s)$, a predictor of the prior action distribution, into its formulation as follows:
\begin{align}\label{eq:puct}
    a^*\!=\argmax_{a \in A(s)} \!\left(\!Q(s,a) \!+\! w \!\cdot\! \pi(a|s) \frac{\sqrt{{N(s)}}}{N(s,a)+1} \!\right) .
\end{align}
The predictor $\pi(a|s)$ can be defined as the sequence probability $p_\theta(a | s, m)$ in the context of LLMs.
A more detailed description of the operators used in MCTS can be found in Appendix~\ref{append:mcts}.


%% file: contents/method.tex
\section{Method}\label{sec:method}

\begin{figure*}[ht!]
     \centering
     \includegraphics[width=\textwidth]{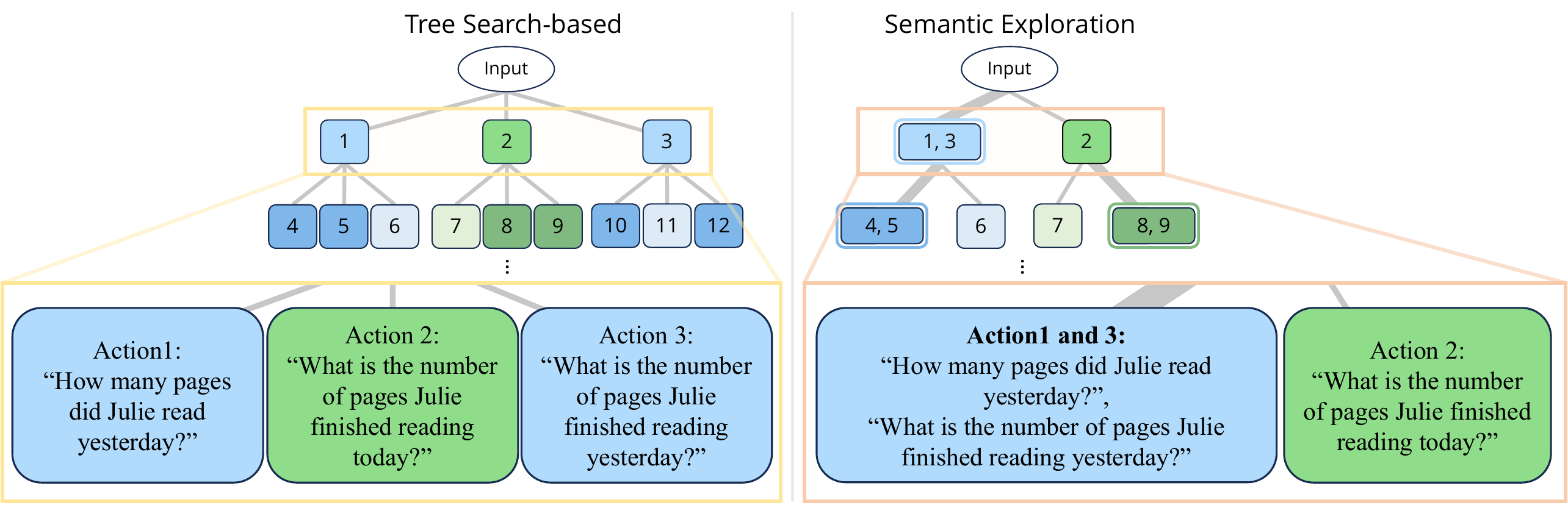}
     \vspace{-7mm}
    \caption{
    Illustration of tree search-based reasoning methods~\cite{yao2023tree, hao2023reasoning} and SE (Ours).
    Nodes with similar color tones (e.g., blue and green) represent semantically similar node, and the thickness of each line indicates the magnitude of the exploration weight.
    }\label{fig:SE}
    \vspace{-1em}
\end{figure*}

Our method consists of adaptive gating (Section~\ref{sec:method:ag}), semantic exploration (Section~\ref{sec:method:se}), and early stopping with weighted aggregation (Section~\ref{sec:method:es}), as shown in Figure~\ref{fig:pipeline}. A detailed description of the algorithm is provided in Appendix~\ref{append:algo}.

\subsection{Adaptive Gating (AG)}\label{sec:method:ag}

AG adaptively determines when to expand the search tree based on the confidence of generated answers.
Specifically, $k$ answers are sampled using a single-path method, such as CoT, and the confidence of these answers is used to determine whether to initiate tree search.
The key insight is that not all problems require the same level of complexity in reasoning. 

First, we generate $k$ reasoning paths using CoT, each producing an output $y^i \sim p_{\theta}(y|x, z_{\le l}^i), \forall i\in[k]$, \textit{i.e.}, corresponding to CoT-SC.
Let $\mathcal{Y}$ denote the set of possible candidate outputs across $k$ reasoning paths.
We define $q(y)$ as the estimated probability of output $y \in \mathcal{Y}$, computed as:
\begin{align}
    q(y)=\frac{1}{k} \sum_{i=1}^{k}\mathbb{I}(y=y^i) \;,
\end{align}
where $\mathbb{I}$ is the indicator function.
Notice that semantical clustering in Section~\ref{sec:method:se} can be used  when calculating $q(y)$, while the indicator function handles cases where the final answer is numeric or discrete.
We use entropy $H(y)$ as a gating function across the $k$ reasoning paths, calculated as:
\begin{align}\label{eq:ent}
    H(y)=-\sum_{y\in \mathcal{Y}}q(y)\log q(y) \;,
\end{align}
which reflects the uncertainty in the model's reasoning.
Based on this, if $H(y)$ is smaller than a predefined threshold $\tau$ indicating high confidence, the final answer $y^{*}$ is determined through majority voting as follows:
\begin{align}
    y^{*} = \arg \max_{y \in \mathcal{Y}} \sum_{i=1}^{k} \mathbb{I}(y = y^i), \;\; 
    \text{if } H(y) \leq \tau \;.
\end{align}
Otherwise, if $H(y)>\tau$,
 indicating lower confidence, we proceed by constructing a search tree and employing our proposed method, Semantic Exploration, as described in the following section. An analysis of entropy distribution is shown in Figure~\ref{fig:ent} in Section~\ref{sec:exp:analysis:ag}.


\subsection{Semantic Exploration (SE)}\label{sec:method:se}
For tree based search, we propose SE, which leverages linguistic semantics to avoid exploring semantically similar nodes. For example, as illustrated in Figure~\ref{fig:SE}, node 1 (``\texttt{How many pages did Julie read yesterday?}'') and node 3 (``\texttt{What is the number of pages Julie finished reading yesterday?}'') convey the identical meaning. The key components of SE are semantic clustering and semantic PUCT.

\paragraph{Semantic clustering}

Semantic clustering groups generated actions based on their semantic equivalence.
At the current node $s$, the tree generates $d$ candidate actions, $A(s)$, using a language model. 

For each pair of actions $(a, a')$ in $A(s)$, the equivalence relation $E(a, a')$ is established by checking for bi-directional entailment between the actions~\cite{kuhn2023semantic, farquhar2024detecting}. 
Specifically, each entailment decision is determined by applying an $\arg\max$ function to the outputs of the DeBERTa-large model~\cite{he2020deberta}, a relatively lightweight model compared to larger LLMs, which is used for single-sentence inputs.
Actions with equivalent meanings are grouped into semantic equivalence clusters $\mathcal{C} = \{C_1, C_2, \dots, C_{d'}\}$, where each cluster $C_i \subseteq A(s)$ contains semantically identical actions and $d'$ represents the total number of clusters.
As illustrated in Figure \ref{fig:SE}, semantic clustering prevents the redundant expansion of subsequent sub-trees from nodes containing actions with equivalent meanings.
Our analysis of semantically unique actions is presented in Section~\ref{sec:exp:analysis:ua}.

\paragraph{Semantic PUCT}
To achieve efficient exploration with semantic clusters $\mathcal{C}$, we propose semantic PUCT. Previous research in LLMs has shown that more frequently generated samples indicate higher significance~\cite{wang2023self}. 
Leveraging this self-consistency principle, we define the predictor of each semantic cluster \(C_i\) as the total probability assigned to the cluster:
\begin{align} 
\pi(C_i|s) = \sum_{a \in C_i} p_\theta(a | s, m) \;.
\end{align}
We extend the PUCT algorithm (Equation~\eqref{eq:puct}) to operate at the level of semantic clusters as follows:
\begin{align} \label{eq:sPUCT} 
C^* \!\!= \! \argmax_{C \in \mathcal{C}} \!\left( \!Q(s,C) \!+\! w \! \cdot \! \pi(C|s) \frac{\sqrt{N(s)}}{N(s,C)+1} \! \right)\!\!\;.
\end{align}
Once a semantic cluster $C_i$ is selected, the highest-probability action within the cluster, \textit{i.e.,} $a^* = \arg\max_{a \in C_i} p_\theta(a \mid s, m)$, is used to construct the prompt for expansion.
We present the effect of semantic PUCT in an ablation study in Section~\ref{sec:exp:ablation:spuct}.





\input{tables/main_instruct}

\subsection{Early Stopping}\label{sec:method:es}

We introduce an early stopping phase in MCTS to decide whether to terminate iterations early based on the confidence of the most probable answer.
To measure this confidence, we use weighted aggregation rewards, which account for the semantic importance of nodes in the search tree.

Let $\mathcal{T}$ denote the set of terminal nodes, and $P(n_j)$ be the set of nodes along the path from the root node to terminal node $n_j \in \mathcal{T}$.
The reward of a terminal node $n_j$ is computed by weighting the size of the semantic cluster $|C(n)|$, where $C(n)$ is the cluster including the node $n$:
\begin{align}
    \label{eq:weighted_aggregation}
    R(n_j) = \sum_{n \in P(n_j)} \lvert C(n) \rvert \cdot r(n), \;\; \forall n_j \in \mathcal{T} \;.
\end{align}
We define $Y(n_j)$ as the extracted answer of a terminal node $n_j$, and $\mathcal{Y'}$ as the set of all extracted answers. 
The aggregated reward $R_{\text{agg}}(y)$ for each answer $y \in \mathcal{Y'}$ is then computed by summing the rewards of all terminal nodes that produce the same answer $y$:
\begin{align}
R_{\text{agg}}(y) = \sum_{n_j \in \mathcal{T}, Y(n_j) = y} R(n_j) \;.
\end{align}
This weighted aggregation ensures that nodes in larger semantic clusters, \textit{i.e.}, more significant nodes, have a greater influence on the reward aggregation.

At each iteration, when a terminal node is reached, the algorithm terminates early if the highest aggregated reward satisfies sufficient confidence threshold $\alpha$:
\begin{align} y^{*} = \arg \max_{y \in \mathcal{Y'}} R_{\text{agg}}(y), \;\; \text{if } \max_{y \in \mathcal{Y'}} R_{\text{agg}}(y) \geq \alpha \;. \end{align}
An ablation study of $\alpha$ is presented in Figure~\ref{fig:token_usage}.

%% file: tables/main_instruct.tex
\begin{table*}[tb!]
\centering
\begin{adjustbox}{width=1.0\textwidth}
\begin{tabular}{cccccccc}
\toprule 
\multicolumn{1}{c}{\multirow{2.5}{*}{Benchmark}} & \multicolumn{1}{c}{\multirow{2.5}{*}{Method}} & \multicolumn{2}{c}{Llama3-8B-Instruct} & \multicolumn{2}{c}{Llama2-13B-Chat} & \multicolumn{2}{c}{Mistral-7B-Instruct} \\
\cmidrule{3-8}
& & Accuracy $\uparrow$ &  \# of inferences $\downarrow$  & Accuracy $\uparrow$ & \# of inferences $\downarrow$ & Accuracy $\uparrow$ & \# of inferences $\downarrow$ \\ 
\midrule
\multirow{6}{*}{GSM8K} & CoT & 0.762	& 1	& 0.330	& 1	& 0.502	& 1 \\
& CoT-SC & 0.845 & 10	& 0.417	& 10	& 0.665	& 10 \\
& ToT & 0.785	& 104.80	& 0.378 & 102.06 & 0.613 & 122.16	  \\
& RAP & 0.825	& 128.40	& 0.372	& 121.69	& 0.667	& 161.86 \\
& \cellcolor[HTML]{EFEFEF}SE (Ours) & \cellcolor[HTML]{EFEFEF}0.850	& \cellcolor[HTML]{EFEFEF}82.63	& \cellcolor[HTML]{EFEFEF}0.403	& \cellcolor[HTML]{EFEFEF}69.47	& \cellcolor[HTML]{EFEFEF}0.675	& \cellcolor[HTML]{EFEFEF}98.79	 \\
& \cellcolor[HTML]{EFEFEF}SEAG (Ours) & \cellcolor[HTML]{EFEFEF}\textbf{0.860}	& \cellcolor[HTML]{EFEFEF}41.69	& \cellcolor[HTML]{EFEFEF}\textbf{0.435}	& \cellcolor[HTML]{EFEFEF}53.09	& \cellcolor[HTML]{EFEFEF}\textbf{0.685}	& \cellcolor[HTML]{EFEFEF}84.14 \\

\midrule
\multirow{6}{*}{ARC} & CoT & 0.818 & 1 & 0.598  & 1 & 0.688 & 1 \\
& CoT-SC & 0.823 & 10 & 0.637 & 10 & 0.708 & 10 \\
& ToT & 0.797 & 149.59 & 0.590 & 209.05 & 0.615 & 221.36 \\
& RAP & 0.812 & 196.96 & 0.637 & 247.22  & 0.705 & 313.20 \\
& \cellcolor[HTML]{EFEFEF}SE (Ours) & \cellcolor[HTML]{EFEFEF}0.830 & \cellcolor[HTML]{EFEFEF}96.32 & \cellcolor[HTML]{EFEFEF}0.632 & \cellcolor[HTML]{EFEFEF}146.20 & \cellcolor[HTML]{EFEFEF} 0.713 & \cellcolor[HTML]{EFEFEF}155.13 \\
& \cellcolor[HTML]{EFEFEF}SEAG (Ours) & \cellcolor[HTML]{EFEFEF}\textbf{0.848} & \cellcolor[HTML]{EFEFEF}46.15 & \cellcolor[HTML]{EFEFEF}\textbf{0.638} & \cellcolor[HTML]{EFEFEF}26.62 & \cellcolor[HTML]{EFEFEF}\textbf{0.725} & \cellcolor[HTML]{EFEFEF}110.95 \\


\bottomrule
\end{tabular}
\end{adjustbox}
\caption{
Comparison of reasoning methods effectiveness in accuracy and efficiency in the number of inferences for both GSM8K and ARC datasets. \textbf{Bold texts} indicate the best accuracies in each setting.
}\label{tab:main}
\vspace{-.5em}
\end{table*}

%% file: contents/experiment.tex
\begin{figure*}[htb!]
     \centering
     \includegraphics[width=0.95\textwidth]{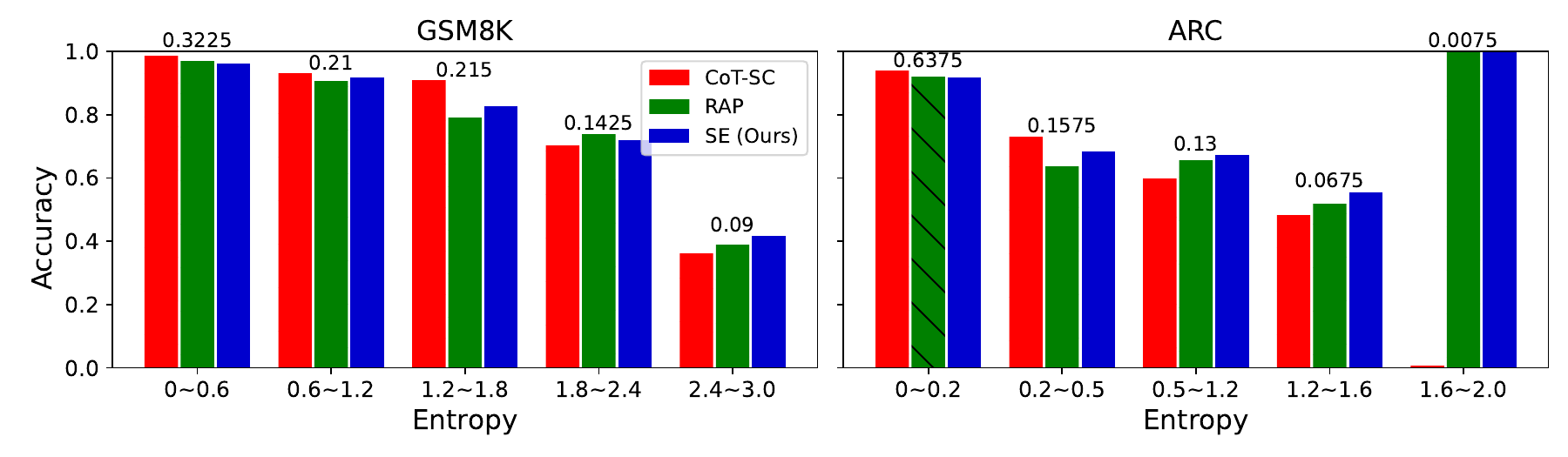}
     \vspace{-3mm}
    \caption{Accuracy comparison of CoT-SC, RAP, and SE (Ours) methods across different entropy ranges for GSM8K and ARC datasets.
    The numbers at the top of each entropy bin represent the proportion of problems within the corresponding entropy range.
    }\label{fig:ent}
\end{figure*}

\section{Experiments}
In this section, we present the experimental evaluation of SEAG. Section~\ref{sec:exp:setup} describes the experimental setup, including datasets and models. The main evaluation results are presented in Section~\ref{sec:exp:result}. Detailed analyses follows in Section~\ref{sec:exp:analysis}, along with ablation studies in Section~\ref{sec:exp:ablation}. Additional experimental details are provided in Appendix~\ref{append:exp}.


\subsection{Experimental Setup} \label{sec:exp:setup}
\paragraph{Baselines} 
We consider four reasoning methods as baselines across sequential and tree-based reasoning approaches.
CoT~\cite{wei2022chain} and CoT-SC~\cite{wang2023self} are sequential reasoning approaches, where CoT generates step-by-step reasoning paths, and CoT-SC extends this by incorporating self-consistency.
ToT~\cite{yao2023tree} and RAP~\cite{hao2023reasoning} use tree search algorithms to explore multiple reasoning paths, utilizing algorithms such as beam search and MCTS, respectively.
For SEAG, SE, and RAP, the number of MCTS iterations is 10 and the number of actions is 4.
For AG, we set $k=10$ across all experiments.
To ensure a fair comparison, we use identical prompts and the same number of in-context examples across all methods. The primary difference lies in the specific search mechanisms employed by each method.
We present the detailed prompts for each baseline in Appendix~\ref{append:prompts}.

\paragraph{Datasets} 
For evaluating the reasoning ability, we use two standard reasoning tasks:
GSM8K~\cite{cobbe2021gsm8k}, a dataset of 8.5k math word problems requiring multi-step numerical reasoning, and
AI2 Reasoning Challenge (ARC)~\cite{clark2018think}, a dataset containing 7.8k multiple-choice science questions sourced from grade-level science exams.
For evaluation, we randomly sample 400 instances from the test sets of GSM8K and ARC.

\paragraph{Models} We conduct experiments with different LLMs using Llama3-8B-Instruct~\cite{dubey2024llama3}, Llama2-13B-Chat~\cite{touvron2023llama2}, and Mistral-7B-Instruct-v0.3~\cite{jiang2023mistral} to demonstrate the robustness of SEAG.

\paragraph{Evaluation metrics}
We use accuracy and computational costs, which are measured in terms of the number of inferences and tokens. The number of inferences refer to the total LLM calls required during the reasoning process per instance. 
Input tokens and output tokens denote the total number of tokens processed during reasoning and generated by the model as output, respectively, for each instance. 

\subsection{Main Results}\label{sec:exp:result}
Table~\ref{tab:main} compares the accuracy and inference efficiency of six reasoning methods evaluated on the GSM8K and ARC using three different LLMs. The results for the base LLM are presented in in Appendix~\ref{append:exp:results}.
SEAG consistently outperforms all baseline methods in accuracy across all datasets and models, demonstrating its robustness and effectiveness. 
Notably, compared to RAP, which is our closest baseline, SEAG achieves a 4.3\% increase in accuracy while requiring only 31\% as many inferences as RAP, on average. This highlights SEAG's ability to deliver superior reasoning performance with improved computational efficiency. 
Further discussions on inference cost in terms of tokens are provided in Appendix~\ref{append:exp:results}.

When evaluating SE independently, SE also achieves higher accuracy and fewer inference costs compared to tree search-based methods, RAP and ToT, as illustrated in Figure~\ref{fig:scatter:llama3}. By incorporating AG, SEAG further improves performance by adaptively determining reasoning paths. This allows SEAG to capitalize on CoT-SC’s strength in internal reasoning, achieving higher accuracy while also enhancing computational efficiency. Additional plots using token cost as an alternative cost metric are presented in Appendix~\ref{append:scatter}.


\subsection{Analyses}\label{sec:exp:analysis}

\paragraph{Necessity of AG}
\label{sec:exp:analysis:ag}

Using the results of 10 independent CoT-SC samplings, we calculate confidence in terms of entropy values as defined in Equation~\eqref{eq:ent}. Problems are categorized into several groups based on their entropy values. Figure~\ref{fig:ent} presents the results of applying CoT, RAP, and SE to problems within each group.
Notice that low entropy indicates that the model generates the same answer with confidence, while high entropy suggests that the model provides different answers across different samplings. We observe that CoT-SC achieves high accuracy for problems with low entropy. In contrast, RAP and SE, which incorporate tree search, demonstrate superior accuracy for high-entropy problems. 

\paragraph{Efficiency improvement by SE}
\label{sec:exp:analysis:ua}
\input{tables/node_counts}
To evaluate the impact of SE, we measure the number of unique semantic clusters $d'$ (\textit{i.e.}, the number of nodes expanded after applying semantic clustering) and the reduction rate at each depth when four actions, $d=4$, are generated in each state.
Table~\ref{tab:node_counts} presents that semantic clustering effectively reduces the redundant search space from 25-45\% at depth 1 to 50-65\% at depth 4. In overall, semantic clustering eliminates approximately 20–60\% of semantically redundant paths, significantly reducing the search space across different depths and datasets. These results indicate that SE avoids repeatedly expanding and exploring semantically redundant paths by incorporating semantic equivalence. Additionally, in Appendix \ref{append:discussion}, we provide a brief analysis on prompt design modification to encourage diverse action sampling as an alternative approach to SE.

\begin{figure}[t!]
     \centering
     \begin{subfigure}[b]{\columnwidth}
         \centering
         \includegraphics[width=.98\columnwidth]{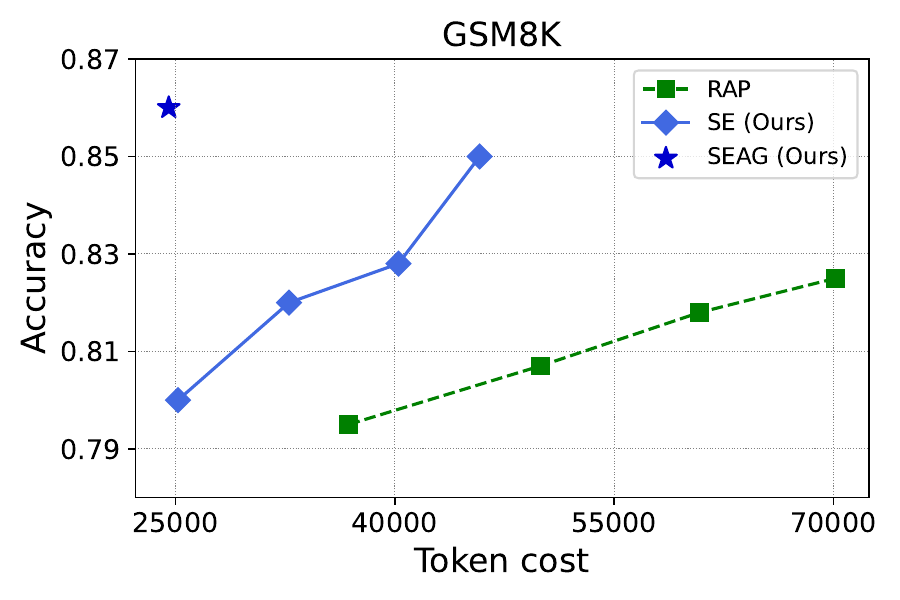}
         \label{fig:tok:1}
     \end{subfigure}
     \begin{subfigure}[b]{\columnwidth}
         \centering
         \includegraphics[width=.98\columnwidth]{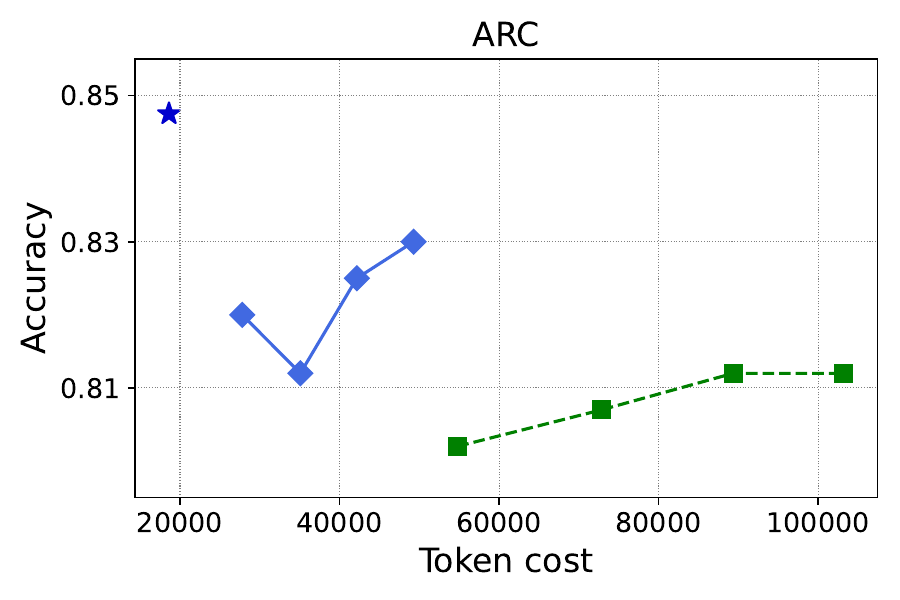}
         \label{fig:tok:2}
     \end{subfigure}
     \caption{Comparisons of reasoning methods in average accuracy with increasing token cost.
     The token cost varies along with the different hyperparameters such as the number of MCTS iterations $k$ and early stopping thresholds $\alpha$.
     }
     \label{fig:token_usage}
\end{figure}

\paragraph{Effect of the number of tokens}
\label{sec:exp:ablation:tu}
We analyze the effect of increasing average token cost on the accuracy improvement of the methods in Figure~\ref{fig:token_usage}. Token cost is calculated based on the pricing policy of commercial LLM APIs, where output tokens are assigned a cost approximately four times higher than input tokens. For RAP, token cost is controlled by varying the number of MCTS iterations, set to 4, 6, 8, and 10. For SE, we adjusts the early stopping threshold $\alpha$, set to 5, 7, 9, and 11, while fixing the number of MCTS iterations at 10. As shown in Figure~\ref{fig:token_usage}, SE demonstrates a steeper performance improvement compared to RAP. Furthermore, by incorporating AG into SE, SEAG shows the highest accuracy (when $\alpha=11$), and achieves further improvements in both accuracy and token cost.

\paragraph{Latency analysis}
\input{tables/average-latency}
We assess the end-to-end latency of each reasoning method on 100 randomly selected ARC instances using the Llama3-8B-Instruct on a single NVIDIA RTX A5000 GPU. 
As shown in Table~\ref{tab:latency_comparison}, SE achieves a reduced average latency of 76.44 seconds, including only 2.54 seconds for semantic clustering, compared to ToT (120.63 s) and RAP (151.19 s). 
Our final method SEAG further improves computational efficiency by adaptively invoking SE only when necessary. Specifically, SEAG terminates early after CoT-SC in approximately 67\% of cases, resulting in a low average latency of $\approx$13.67 seconds, and utilizes the full SE reasoning process in the remaining 33\% of cases with an average latency of $\approx$90.11 seconds.
The ordering and trend of latency across methods are consistent with those observed in LLM inference counts and token usage.

\subsection{Ablation Studies}\label{sec:exp:ablation}


\input{tables/puct_ablation}

\input{tables/aggregation_ablation}

\paragraph{Improvement by semantic PUCT}
\label{sec:exp:ablation:spuct}
To evaluate the effectiveness of the proposed action selection algorithm, semantic PUCT in equation~\eqref{eq:sPUCT}, we conduct an ablation study with two existing algorithms: UCT, which relies solely on a count-based uncertainty term in equation~\eqref{eq:puct}, and PUCT, which incorporates an uncertainty term involving $\pi(a|s)$ and a count-based term in equation~\eqref{eq:puct}. In Table~\ref{tab:aggregation_ablation}, we demonstrate that Semantic PUCT outperforms the existing algorithms while maintaining comparable efficiency by encouraging exploration of semantically probable clusters with $\pi(C|s)$.

\paragraph{Improvement by weighted aggregation of rewards}
\label{sec:exp:ablation:wa}
To evaluate the effectiveness of the aggregation strategy weighted by $|C(n)|$, we conducted a comparison with an aggregation strategy where all actions were weighted equally (\textit{i.e.}, manually set $|C(n)|$ to 1 in equation~\ref{eq:weighted_aggregation}), regardless of clustering. The experiments were conducted using the SEAG method. To ensure a fair comparison of LLM computation costs, $\alpha$ was set to 11 for the weighted aggregation and 8 for the equal weighting strategy. As shown in Table~\ref{tab:aggregation_ablation}, the weighted aggregation consistently achieved higher accuracy across both benchmarks while incurring lower LLM computation costs.

%% file: tables/node_counts.tex

\begin{table}[tb!]
\centering
\begin{adjustbox}{width=\linewidth}
\begin{tabular}{ccccc}
\toprule 
  & \multicolumn{2}{c}{GSM8K} & \multicolumn{2}{c}{ARC} \\
\cmidrule{2-5}
Depth & \makecell{\# of \\ semantic \\ clusters} & \makecell{Reduction \\ rate (\%)}  & \makecell{\# of \\ semantic \\ clusters} & \makecell{Reduction \\ rate (\%)} \\ 
\midrule
1 & 2.31	& 42.36	& 3.00	& 25.00 \\
2 & 2.15	& 46.33 & 3.22  & 19.38	  \\
3 & 2.07	& 48.34	& 2.60	& 35.02	\\
4 & 1.81	& 54.70	& 1.55	& 61.25	\\

\bottomrule
\end{tabular}
\end{adjustbox}
\caption{
The average number of semantic clusters (counts) and reduction rate (\%) at each depth when four actions are generated at every step, $d=4$ for both GSM8K and ARC datasets. 
} \label{tab:node_counts}
\end{table}

%% file: tables/average-latency.tex
\begin{table}[t!]
\centering
\begin{adjustbox}{width=0.7\linewidth}
\begin{tabular}{cc}
\toprule
Method & Average latency (sec) \\
\midrule
CoT & 1.43 \\
CoT-SC & 13.67 \\
ToT & 120.63 \\
RAP & 151.19 \\
\cellcolor[HTML]{EFEFEF}SE (Ours) & \cellcolor[HTML]{EFEFEF}76.44 \\
\cellcolor[HTML]{EFEFEF}SEAG (Ours) & \cellcolor[HTML]{EFEFEF}38.89 \\
\bottomrule
\end{tabular}
\end{adjustbox}
\caption{Average end-to-end latency of reasoning methods evaluated on the ARC dataset using Llama3-8B-Instruct.}
\vspace{-1em}
\label{tab:latency_comparison}
\end{table}

%% file: tables/puct_ablation.tex
\begin{table}[t!]
\centering
\begin{adjustbox}{width=\linewidth}
\begin{tabular}{cccccc}
\toprule



& Action selection method & Accuracy & \# of inferences  \\ 

\midrule
\multirow{3}{*}{GSM8K} & UCT & 0.828 & 83.53 \\
 & PUCT & 0.840 & 85.39  \\
& \cellcolor[HTML]{EFEFEF}Semantic PUCT (Ours) & \cellcolor[HTML]{EFEFEF}\textbf{0.850} & \cellcolor[HTML]{EFEFEF}82.63 \\

\midrule
\multirow{3}{*}{ARC} & UCT & 0.815 & 95.16 \\
 & PUCT & 0.818 & 97.88 \\
& \cellcolor[HTML]{EFEFEF}Semantic PUCT (Ours) & \cellcolor[HTML]{EFEFEF}\textbf{0.830} & \cellcolor[HTML]{EFEFEF}96.32 \\

\bottomrule
\end{tabular}
\end{adjustbox}
\caption{Ablation study of action selection algorithms, including UCT, PUCT, and Semantic PUCT (Ours), presenting average accuracy and the number of inferences for both GSM8K and ARC datasets.} \label{tab:puct_ablation}
\end{table}

%% file: tables/aggregation_ablation.tex
\begin{table}[t!]
\centering
\begin{adjustbox}{width=\linewidth}
\begin{tabular}{cccccc}
\toprule



& Aggregation method & Accuracy & \# of inferences  \\ 

\midrule
\multirow{2}{*}{GSM8K} & Equal weighting & 0.845 & 95.89 \\
& \cellcolor[HTML]{EFEFEF}Weighting by $|C|$ & \cellcolor[HTML]{EFEFEF}\textbf{0.850} & \cellcolor[HTML]{EFEFEF}82.63 \\

\midrule
\multirow{2}{*}{ARC} & Equal weighting & 0.828 & 111.00 \\
& \cellcolor[HTML]{EFEFEF}Weighting by $|C|$ & \cellcolor[HTML]{EFEFEF}\textbf{0.830} & \cellcolor[HTML]{EFEFEF}96.32 \\

\bottomrule
\end{tabular}
\end{adjustbox}
\caption{Ablation study of aggregation methods, including equal weighting and weighted aggregation (Ours), presenting average accuracy and the number of inferences for both GSM8K and ARC datasets.} \label{tab:aggregation_ablation}
\end{table}

%% file: contents/conclusion.tex
\section{Conclusion}

In this paper, we propose Semantic Exploration with Adaptive Gating (SEAG), a framework that enhances the efficiency and accuracy of multi-step reasoning tasks in LLMs. SEAG addresses two key issues in tree-based reasoning methods: (i) the unnecessary use of complex reasoning techniques for tasks solvable with simpler approaches, and (ii) the generation of semantically redundant nodes during exploration. By combining adaptive gating to evaluate task complexity, semantic exploration to minimize redundancy, and early stopping to reduce computational overhead, SEAG achieves significant performance improvements.

%% file: contents/limitation.tex
\section*{Limitations}

Our method focuses on leveraging only internal knowledge to enhance the efficiency of tree search-based reasoning methods. This focus ensures a fair evaluation of the method's inherent efficiency without reliance on external tools or feedback. Expanding the approach to incorporate such external resources could further improve the reasoning process, presenting a promising direction for future work.
Furthermore, the experiments are primarily conducted on benchmarks where final answers are provided in a discrete form, rather than free-form natural language generation. Evaluating the method on tasks requiring open-ended responses is left as future work.

%% file: contents/ack.tex
\section*{Acknowledgements}

This work was supported by Institute of Information \& communications Technology Planning \& Evaluation (IITP) grants funded by the Korea government (MSIT) (No.RS-2019-II191906, Artificial Intelligence Graduate School Program (POSTECH); No.RS-2024-00457882, AI Research Hub Project; No.RS-2024-00509258, Global AI Frontier Lab) and the Korea Institute for Advancement of Technology (KAIT), funded by the Ministry of Trade, Industry and Energy (MOTIE), Republic of Korea (No.RS-2025-00564342).
\vspace*{.8cm}

%% file: contents/appendix.tex
\section*{Appendix}\label{sec:appendix}

\renewcommand{\thefigure}{A\arabic{figure}}
\setcounter{figure}{0}

\renewcommand{\thetable}{A\arabic{table}}
\setcounter{table}{0}

\begin{table*}[htp!]
    \centering
    \begin{tabular}{ccc}
        \toprule
        Setting & Parameter & Value \\
        \midrule
         \multirow{3}{*}{LLM-related setting} &  temperature & $0.8$ \\
         & top-$k$  & $50$ \\
         & top-$p$ & $0.95$ \\
         \midrule
         \multirow{2}{*}{General setting } & batch size & $1$ \\
         & the number of examples (prompts) & $1$ \\
         \midrule
         \multirow{6}{*}{MCTS setting} & depth limit & $5$ \\
         & the number of iterations & $10$ \\
         & the number of actions & $4$ \\
         & reward alpha & $0.5$ \\
         & the number of confidences & $8$ \\
         & a default value of reward confidence  & $0.8$ \\
         \midrule
         \multirow{2}{*}{ToT setting} & beam size & 3 \\
         & depth limit & $5$ \\
         \midrule
         CoT-SC setting & the number of self-consistency & 10 \\ 
         \bottomrule
    \end{tabular}
    \caption{Default hyperparameters for SE, RAP, ToT, and CoT-SC. Parameters are grouped into LLM-related, general, and method-specific settings.}
    \label{tab:appendix_hyperparam}
\end{table*}

\section{Further Details of Reasoning Frameworks in Language Models}\label{append:reasoning}

    \paragraph{Chain-of-Thought (CoT) prompting} \cite{wei2022chain} sequentially generates thoughts $z_1, \dots, z_l$, where  $z_i \sim p_{\theta}(z_i|x,z_{<i})$, and then generates the output with these thoughts, i.e., $y \sim p_{\theta}(y|x,z_{\le l})$.
    
    \paragraph{Self-Consistency with CoT (CoT-SC)} \cite{wang2023self}.
    CoT-SC is an ensemble-based method that leverages $k$ independently sampled reasoning paths, each generated using CoT prompting. 
    Specifically, for each $i \in [k]$, a reasoning path generates an output $y^i \sim p_{\theta}(y|x,z^{i}_{\le l})$. The final output is determined by majority voting across $k$ paths: $\argmax_{y\in\mathcal{Y}}\sum_{i=1}^k\mathbb{I}(y^i=y)$, where $\mathcal{Y}$ is the set of all possible candidate outputs and $\mathbb{I}(\cdot)$ is the indicator function.

    \paragraph{Tree-of-Thought (ToT) prompting} \cite{yao2023tree}.
    For exploring multiple reasoning paths, ToT prompting extends CoT by structuring the reasoning process as a tree structure, where each node represents a reasoning state $s = [x, z_1, ..., z_i]$ with input $x$ and the sequence of intermediate thoughts.
    Intermediate thoughts are generated sequentially as $z_i \sim p_{\theta}(z_i|x, z_{<i})$.
    The exploration relies on search algorithms, such as Breath-First Search (BFS) or Depth-First Search (DFS), to evaluate and prioritize paths using heuristics based on LM evaluations of each state.

\section{Reward Design in MDP}\label{append:reward}
    We adapt the reward design proposed in RAP \cite{hao2023reasoning}, focuses on evaluating the feasibility and desirability of reasoning steps.
    This approach incorporates a reward function 
    $r_t=r(s_t, a_t)\in \mathbb{R}$ to assess the impact of an action $a_t$ to a state $s_t$.
    Our reward design integrates two key components: self-evaluation of the action's helpfulness and confidence of the resulting state.
    
    \paragraph{Self-evaluation by the LLM}
    The model can self-assess the helpfulness of reasoning by answering questions, such as “Is this reasoning step useful?”. The reward is derived from the probability of the token “Yes”, which provides an estimate of the LLM's confidence in the correctness of its reasoning. This self-evaluation mechanism can adapt to task-specific nuances.
    
    \paragraph{Confidence of the state} 
    The confidence of a state is determined by sampling multiple answers from the model and calculating the proportion of the most frequent answer. A higher confidence score indicates stronger alignment with the LLM’s internal knowledge, promoting more reliable reasoning steps.
    For ARC dataset, which requires answers in natural language, we also incorporate the concept of semantic exploration to merge similar states, enabling more accurate confidence estimation. In SE, we construct the LLM prompt by uniformly sampling actions from a semantic cluster when generating multiple answers.

\section{Monte Carlo Tree Search (MCTS)}\label{append:mcts}
    \paragraph{Selection} In the selection phase, algorithm chooses a move at a node or an action in the tree. The choice is based on pre-defined strategies, the typically the Upper Confidence Bound applied to Trees (UCT) \cite{kocsis2006bandit}, which is based on the upper confidence bound (UCB) \cite{auer2002finite}.
    At node $s$, we select the action with balance exploration trying less visited actions and exploitation choosing actions with higher state-action value function $Q: \mathcal{S} \times \mathcal{A} \mapsto \mathbb{R}$, which estimated the expected future reward with $(s, a)$.
    
    \begin{align}
        a^*=\argmax_{a \in A(s)} \left(Q(s,a)+w \sqrt{\frac{\ln{N(s)}}{N(s,a)}} \right) \;,
    \end{align}
    where $N(\cdot)$ denotes the total number of visits in previous iterations, $A(s)$ is the set of possible actions at state $s$, and $w$ is the constant of balancing exploration and exploitation. 
    
    \paragraph{Expansion} Expansion corresponds to generate new child nodes in a decision tree.
    Given the state at the leaf node, the LM samples $d$ possible actions, forming the action set $A(s)$ which requires $d$ LLM inferences. It then predicts the corresponding next states (the $d$ child nodes). 
    
    \paragraph{Simulation} 
    Simulation corresponds to estimate the $Q$-value at a given node by simulating future states from the node. The candidate actions are evaluated using reward function $r$, selecting the action with the highest local reward $a' = \arg\max_{a \in A(s)}r(s,a)$.
    
    \paragraph{Back-propagation}  
    When  a terminal state is reached, marking the completion of a single reasoning path within one iteration, the $Q$ values of all nodes along the path are updated.
    After completing a $k$ of MCTS iterations, the algorithm terminates, selecting the final answer.
    terminate algorithm and aggregation of lots of reasoning traces.

\section{SEAG Algorithm}\label{append:algo}

Algorithm~\ref{alg:cas} presents the detailed procedure of SEAG, illustrating its key steps: adaptive gating, semantic exploration, and early stopping.

\begin{algorithm*}[htp!]
\caption{Semantic Exploration with Adaptive Gating (SEAG)}\label{alg:cas}
\begin{algorithmic}[1]
    \STATE \textbf{Input:} Input $x$, the number of reasoning paths $k$, $k'$, thresholds $\tau$ (entropy), $\alpha$ (reward confidence), a semantic equivalence relation $E(\cdot, \cdot)$
    \STATE \textbf{Output:} Final answer $y^*$
    
    \STATE Generate $k$ reasoning paths $\{y^i\}_{i=1}^{k}$ using CoT: \hfill \textbf{Step 1: Adaptive Gating} 
    \[
    y^i \sim p_{\theta}(y|x, z_{\le l}^i), \quad \forall i \in [k]
    \]
    \STATE Compute the estimated probability for each $y \in \mathcal{Y}$, where $\mathcal{Y}$ is the set of candidate outputs across $k$ paths:
    $q(y) = \frac{1}{k} \sum_{i=1}^{k} \mathbb{I}(y = y^i)$
    \STATE Compute entropy 
    $H(y) = -\sum_{y \in \mathcal{Y}} q(y) \log(q(y))$
    \IF{$H(y) \leq \tau$}
        \STATE Select final answer using majority voting:
        $y^* = \arg \max_{y \in \mathcal{Y}} \sum_{i=1}^{k} \mathbb{I}(y = y^i)$
        \STATE \textbf{Return:} Final answer $y^*$
    \ELSE
        \FOR{each iteration from $1$ to $k'$}
            \STATE Set current node $s$.
            \REPEAT
                \STATE Generate candidate actions $A(s)$. \hfill \textbf{Step 2: Semantic Exploration}
                \STATE Group actions into semantic clusters $\mathcal{C}=\{C_1, C_2, \dots, C_{d'}\}$ using $E(a, a')$.

                \STATE Compute $\pi(C_i|s) = \sum_{a \in C_i} p_\theta(a | s, m)$ for each semantic cluster $C_i$.
            
            \STATE Select a semantic cluster $C$ via semantic PUCT:
            \[
            C^* \!= \! \argmax_{C \in \mathcal{C}} \!\left( \!Q(s,C) \!+\! w \! \cdot \! \pi(C|s) \frac{\sqrt{N(s)}}{N(s,C)+1} \! \right)\! \;.
            \]

             \STATE Select action $a^* = \argmax_{a \in C^*} p_\theta(a \mid s, m)$ and expand node $s$.
    
            \UNTIL{A terminal node is reached.}
            
            \STATE For terminal nodes $n_j \in \mathcal{T}$, compute path rewards: \hfill \textbf{Step 3-1: Weighted Aggregation}
            \[
            R(n_j) = \sum_{n \in P(n_j)} |C(n)| \cdot r(n)
            \]
            \STATE Define $Y(n_j)$ as the extracted answer of a terminal node $n_j$ and $\mathcal{Y'}$ as the set of answers. 
            \STATE Aggregate rewards for each answer $y \in \mathcal{Y'}$:
            \[
            R_{\text{agg}}(y) = \sum_{n_j \in \mathcal{T}, Y(n_j) = y} R(n_j)
            \]
            \IF{$\max_{y \in \mathcal{Y}} R_{\text{agg}}(y) \geq \alpha$} \hfill \textbf{Step 3-2: Early Stopping}
                \STATE Select final answer: $y^* = \arg \max_{y \in \mathcal{Y}} R_{\text{agg}}(y)$
                \STATE \textbf{Return:} Final answer $y^*$
            \ELSE
                \STATE // \textit{Back-propagation}
            \ENDIF
            \STATE \textbf{Return:} Final answer $y^*$
    \ENDFOR
    \ENDIF
\end{algorithmic}
\end{algorithm*}

\section{Experimental Settings}\label{append:exp}

    We describe the detailed experimental settings to ensure reproducibility.
    Table \ref{tab:appendix_hyperparam} presents the default hyperparameters for each method.

    All experiments were conducted using a single GPU and for Llama3-8B and Mistral-7B models, and two GPUs for Llama2-13B model. The total computational resource is required to produce the results in Table~\ref{tab:main} is approximately 832 GPU hours. We utilize RTX 3090, A5000, and A6000 GPUs for all experiments.
    Due to the high computational cost, we report results based on a single run.

\section{Efficiency with Token Cost}\label{append:scatter}
\begin{figure*}[htb!]
     \centering
     \hfill
     \begin{subfigure}[b]{0.45\textwidth}
         \centering
         \includegraphics[width=\textwidth]{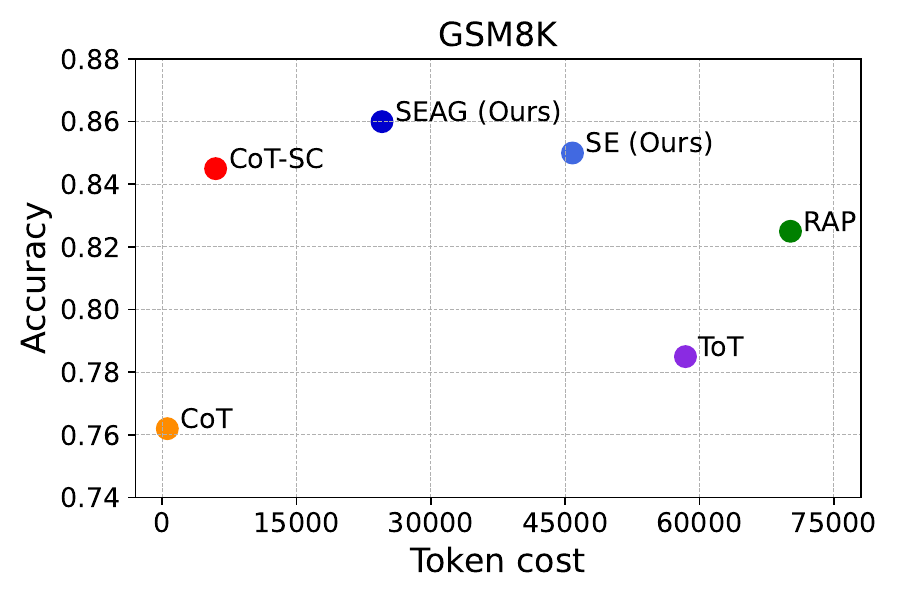}
     \end{subfigure}
     \hfill
     \begin{subfigure}[b]{0.45\textwidth}
         \centering
         \includegraphics[width=\textwidth]{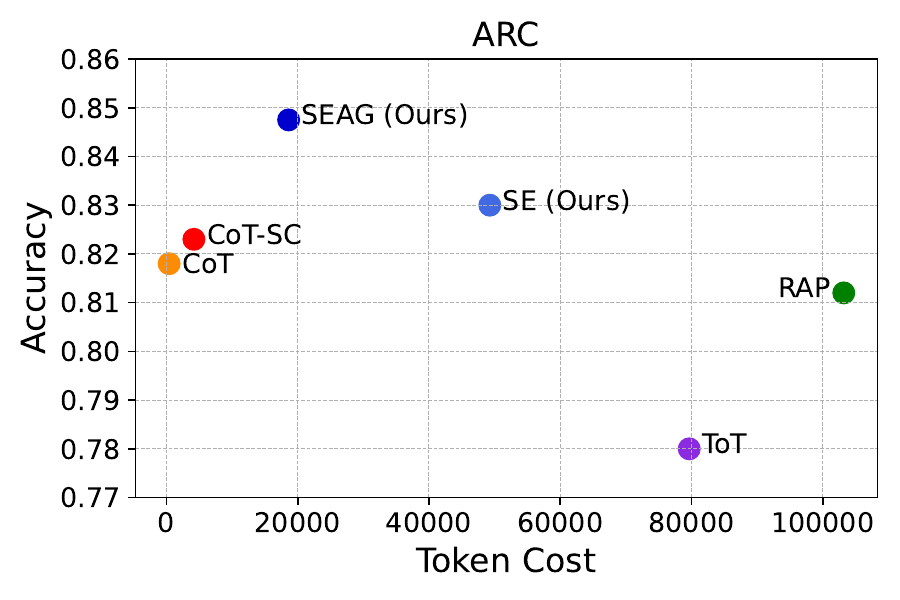}
     \end{subfigure}
     \hfill
        \caption{Scatter plots of accuracy and token cost for baselines and our methods, SE and SEAG, with GSM8K (left) and ARC (right) using Llama3-8B-Instruct model. }
        \label{fig:scatter:llama3:token}
\end{figure*}

\input{tables/main_base}

\input{tables/cost}

We provide additional scatter plots in Figure~\ref{fig:scatter:llama3:token} where token cost is used as the cost metric instead of the number of inferences as shown in Figure~\ref{fig:scatter:llama3}.
The token cost metric represents the cumulative number of tokens processed by the LLM during inference.
The token cost is measured based on a commonly used pricing policy for LLMs, where the total cost is calculated as $\text{(Token cost)}=
\text{(Input tokens)} + 4 \times \text{(Output tokens)}$.
Figure~\ref{fig:scatter:llama3:token} shows similar patterns to those observed in the main paper. This consistency reinforces the robustness of our methods in achieving superior accuracy while minimizing computational expense across different cost perspectives.

\section{Extended Experimental Results}\label{append:exp:results}

In this section, we provide additional experimental results comparing the accuracy and computational costs of various methods on the GSM8K and ARC benchmarks. Specifically, we present results using both Llama3-8B-Base in Table~\ref{tab:cost:llama3:base} and Llama3-8B-Instruct models in Table~\ref{tab:cost:llama3:instruct}.
Both tables highlight the improvements in accuracy and efficiency achieved by our proposed methods, SE (Ours) and SEAG (Ours), across benchmarks. These improvements are evident not only in accuracy across all baselines but also in reductions in the number of inferences, input tokens, and output tokens compared to tree search-based methods.

\section{Prompt Design Modification for Diverse Action Sampling}\label{append:discussion}
\input{tables/sequential_generation_node_counts}

To encourage the generation of semantically unique actions, one can consider sampling actions in a sequential manner rather than i.i.d. sampling using the same prompt. Specifically, we modify the sentence in the original prompt, ``Given a question, please decompose it into sub-questions.'' into ``Given a question, please decompose it into sub-questions \textit{with a distinct meaning from the following sub-questions: `How many pages did Julie read yesterday?', `What is the number of pages Julie finished reading today?'}.'' when sampling the third action to steer the LLM toward generating actions with different meanings. 

Following the procedure in Table \ref{tab:node_counts}, we implement SE using both the i.i.d. and sequential manners of action generation and compare the number of unique semantic clusters expanded with Llama3-8B-Instruct. Interestingly, the sequential manner designed to encourage the generation of semantically distinct actions results in fewer unique semantic clusters compared to the standard i.i.d. sampling as shown in \ref{tab:sequential_generation_node_counts}. Additionally, we observe a performance drop in accuracy with the sequential manner compared to i.i.d. sampling, from 0.850 to 0.812 on GSM8K and from 0.830 to 0.823 on ARC. Thus, while generating non-redundant actions through prompt-level modifications can be an interesting direction, it seems to be non-trivial. Moreover, conditioning action generation on previously sampled actions may shift the sampling distribution, potentially limiting the virtue of semantic consistency inherent in i.i.d. sampling, which is a key advantage used in SE.

\section{Prompt}\label{append:prompts}

    Figure~\ref{fig:appendix-prompt-cot}, \ref{fig:appendix-prompt-gsm}, and \ref{fig:appendix-prompt-arc}  present examples of prompts.
    We use a 1-shot example for all inferences, including prompts for answer, action, and reward generation in  tree search-based reasoning and CoT prompting.

\input{contents/appendix_prompt}

%% file: tables/main_base.tex
\begin{table*}[ht!]
\centering
\begin{adjustbox}{width=.8\textwidth}
\begin{tabular}{cccccc}
\toprule
Benchmark & Method & Accuracy $\uparrow$ & \# of inferences $\downarrow$ & Input tokens $\downarrow$ & Output tokens $\downarrow$ \\ 

\midrule
\multirow{6}{*}{GSM8K} & CoT & 0.445 & 1 & 218.37 &	85.26 \\
 & CoT-SC & 0.598 & 10 & 2183.76 & 906.97 \\
 & ToT & 0.598 & 121.93 & 57187.68  & 2856.04 \\
 & RAP & 0.665 & 146.18 & 67526.58 & 3113.34 \\
 & \cellcolor[HTML]{EFEFEF}SE (Ours) & \cellcolor[HTML]{EFEFEF}0.675 & \cellcolor[HTML]{EFEFEF}92.89 & \cellcolor[HTML]{EFEFEF}42832.03 & \cellcolor[HTML]{EFEFEF}1969.56 \\
& \cellcolor[HTML]{EFEFEF}SEAG (Ours) & \cellcolor[HTML]{EFEFEF}\textbf{0.683} & \cellcolor[HTML]{EFEFEF}64.84 & \cellcolor[HTML]{EFEFEF}25853.65 & \cellcolor[HTML]{EFEFEF}1263.95 \\

\midrule
\multirow{6}{*}{ARC} & CoT & 0.713 & 1 & 215.80 & 29.56 \\
 & CoT-SC & 0.787 & 10 & 2157.93 & 362.51 \\
 & ToT & 0.748 & 169.82 & 75622.06 & 2886.81\\
 & RAP & 0.767 & 208.82 & 91586.37 & 3224.82 \\
& \cellcolor[HTML]{EFEFEF}SE (Ours) & \cellcolor[HTML]{EFEFEF}0.775 & \cellcolor[HTML]{EFEFEF}103.55 & \cellcolor[HTML]{EFEFEF}44404.50 & \cellcolor[HTML]{EFEFEF}1511.65 \\
& \cellcolor[HTML]{EFEFEF}SEAG (Ours) & \cellcolor[HTML]{EFEFEF}\textbf{0.788} & \cellcolor[HTML]{EFEFEF}12.33 & \cellcolor[HTML]{EFEFEF}3155.79 & \cellcolor[HTML]{EFEFEF}397.37\\

\bottomrule
\end{tabular}
\end{adjustbox}
\caption{Comparison of reasoning methods effectiveness in accuracy and efficiency in the number of inferences, input tokens and output tokens for
both GSM8K and ARC datasets using Llama3-8B-Base.} \label{tab:cost:llama3:base}
\end{table*}

%% file: tables/cost.tex
\begin{table*}[ht!]
\centering
\begin{adjustbox}{width=.8\textwidth}
\begin{tabular}{cccccc}
\toprule
Benchmark & Method & Accuracy $\uparrow$ & \# of inferences $\downarrow$ & Input tokens $\downarrow$ & Output tokens $\downarrow$ \\ 

\midrule
\multirow{6}{*}{GSM8K} & CoT & 0.762 &	1 &	218.37 &	85.86\\
 & CoT-SC &  0.845 &	10 &	2183.73 &	943.66\\
 & ToT &  0.785 &	104.80 &	49003.05 &	2355.06\\
 & RAP & 0.825 &	128.40 &	59800.59 &	2587.73\\
 & \cellcolor[HTML]{EFEFEF}SE (Ours) & \cellcolor[HTML]{EFEFEF}0.850 & \cellcolor[HTML]{EFEFEF}82.63 & \cellcolor[HTML]{EFEFEF}38743.45 & \cellcolor[HTML]{EFEFEF}1766.21 \\
& \cellcolor[HTML]{EFEFEF}SEAG (Ours) & \cellcolor[HTML]{EFEFEF}\textbf{0.860} & \cellcolor[HTML]{EFEFEF}41.69 & \cellcolor[HTML]{EFEFEF}17501.68 & \cellcolor[HTML]{EFEFEF}1759.50 \\
\midrule
\multirow{6}{*}{ARC} & CoT & 0.818 & 1 & 215.79 & 48.315\\
 & CoT-SC & 0.823 & 10 & 2157.93 & 505.79\\
 & ToT & 0.797 & 149.59 & 67534.70 & 3033.07\\
 & RAP & 0.812 & 196.96 & 88670.52 & 3636.17\\
& \cellcolor[HTML]{EFEFEF}SE (Ours) & \cellcolor[HTML]{EFEFEF}0.830 & \cellcolor[HTML]{EFEFEF}96.32 & \cellcolor[HTML]{EFEFEF}42507.85 & \cellcolor[HTML]{EFEFEF}1692.92 \\
& \cellcolor[HTML]{EFEFEF}SEAG (Ours) & \cellcolor[HTML]{EFEFEF}\textbf{0.848} & \cellcolor[HTML]{EFEFEF}46.15 & \cellcolor[HTML]{EFEFEF}15993.29 & \cellcolor[HTML]{EFEFEF}655.90 \\
\bottomrule
\end{tabular}
\end{adjustbox}
\caption{Comparison of reasoning methods effectiveness in accuracy and efficiency in the number of inferences, input tokens and output tokens for
both GSM8K and ARC datasets using Llama3-8B-Instruct.} \label{tab:cost:llama3:instruct}
\end{table*}

%% file: tables/sequential_generation_node_counts.tex
\begin{table}[tb!]
\centering
\begin{adjustbox}{width=\linewidth}
\begin{tabular}{ccccc}
\toprule 
  & \multicolumn{2}{c}{GSM8K} & \multicolumn{2}{c}{ARC} \\
\cmidrule{2-5}
Depth & \makecell{i.i.d. \\ sampling} & \makecell{Sequential \\ generation}  & \makecell{i.i.d. \\ sampling} & \makecell{Sequential \\ generation} \\ 
\midrule
1 & 2.31	& 1.37	& 3.00	& 1.72 \\
2 & 2.15	& 1.53 & 3.22  & 2.00	  \\
3 & 2.07	& 1.59	& 2.60	& 2.05	\\
4 & 1.81	& 1.53	& 1.55	& 1.17	\\

\bottomrule
\end{tabular}
\end{adjustbox}
\caption{
The average number of semantic clusters (counts) for i.i.d. sampling and sequential generation at each depth when four actions are generated at every step, $d=4$ for both GSM8K and ARC datasets. 
} \label{tab:sequential_generation_node_counts}
\end{table}

%% file: contents/appendix_prompt.tex
\newtcolorbox{prompt}[1]{
  colback=blue!5,
  colframe=blue!35!black,
  fonttitle=\bfseries,
  title={#1},
  left=2pt,
  right=2pt,
  fontupper=\small,
}

\newcommand{\promptEmphExample}{\it \color{blue!75!black}}

\begin{figure*}[ht!]
\begin{prompt}{A CoT prompt for GSM8K}
Please solve the following question step by step and conclude by saying "The answer is".

{\promptEmphExample
Q: Albert is wondering how much pizza he can eat in one day. He buys 2 large pizzas and 2 small pizzas. A large pizza has 16 slices and a small pizza has 8 slices. If he eats it all, how many pieces does he eat that day?

A: He buys 2 large pizzas, so he has 2 * 16 = 32 slices. He buys 2 small pizzas, so he has 2 * 8 = 16 slices. There are 32 slices from the large pizzas and 16 slices from the small pizzas, so he eats 32 + 16 = 48 pieces that day. The answer is 48.
}

Q: Eliza\'s rate per hour for the first 40 hours she works each week is \$10. She also receives an overtime pay of 1.2 times her regular hourly rate. If Eliza worked for 45 hours this week, how much are her earnings for this week?

A:

\end{prompt}

\begin{prompt}{A CoT prompt for ARC}
Please solve the following question step by step and conclude by saying "The answer is".

{\promptEmphExample
Q: Juan and LaKeisha roll a few objects down a ramp. They want to see which object rolls the farthest. What should they do so they can repeat their investigation? Options: A) Put the objects in groups, B) Change the height of the ramp, C) Choose different objects to roll, D) Record the details of the investigation.

A: To ensure their investigation can be repeated, they need to record detailed information about the setup, such as the objects used, the height of the ramp, and surface conditions. Recording these details allows them to recreate the same conditions for reliable comparisons. The answer is D.
}

Q: Which method is the safest way to watch an eclipse of the Sun? Options: A)  Turn away after two or three minutes. B)  Look at the Sun through a long telescope. C)  Cast an image through a pinhole onto a screen. D)  Blink often until your eyes get used to the light..

A:

\end{prompt}






\vspace{-5mm}
\caption{
    Example prompts for CoT and CoT-SC.
    {\promptEmphExample Italic texts} denote 1-shot examples.
}
\label{fig:appendix-prompt-cot}
\end{figure*}

\begin{figure*}[hp!]
\begin{prompt}{An answer generation prompt for GSM8K}
Given a question, please decompose it into sub-questions. For each sub-question, please answer it in a complete sentence, ending with "The answer is". When the original question is answerable, please start the subquestion with "Now we can answer the question: ".

{
\promptEmphExample

Question 1: Albert is wondering how much pizza he can eat in one day. He buys 2 large pizzas and 2 small pizzas. A large pizza has 16 slices and a small pizza has 8 slices. If he eats it all, how many pieces does he eat that day?

Question 1.1: How many slices are in one large pizza?

Answer 1.1: One large pizza has 16 slices. The answer is 16.

Question 1.2: How many slices are there in total from the large pizzas?

Answer 1.2: He buys 2 large pizzas, so 2 * 16 = 32 slices. The answer is 32.

Question 1.3: How many slices are in one small pizza?

Answer 1.3: One small pizza has 8 slices. The answer is 8.

Question 1.4: How many slices are there in total from the small pizzas?

Answer 1.4: He buys 2 small pizzas, so 2 * 8 = 16 slices. The answer is 16.

Question 1.5: Now we can answer the question: How many pieces does he eat that day?

Answer 1.5: There are 32 slices from the large pizzas and 16 slices from the small pizzas, so he eats 32 + 16 = 48 pieces that day. The answer is 48.
}

Question 2: Josh decides to try flipping a house.  He buys a house for \$80,000 and then puts in \$50,000 in repairs. This increased the value of the house by 150\%.  How much profit did he make?

Question 2.1: How much did Josh spend on the house and repairs in total?

Answer 2.1:

\end{prompt}

\begin{prompt}{An action generation prompt for GSM8K}

Given a question, please decompose it into sub-questions. For each sub-question, please answer it in a complete sentence, ending with "The answer is". When the original question is answerable, please start the subquestion with "Now we can answer the question: ".

{ \promptEmphExample

Question 1: Albert is wondering how much pizza he can eat in one day. He buys 2

large pizzas and 2 small pizzas. A large pizza has 16 slices and a small pizza has 8 slices. If he eats it all, how

many pieces does he eat that day?

Question 1.1: How many slices are in one large pizza?

Answer 1.1: One large pizza has 16 slices. The answer is 16.

Question 1.2: How many slices are there in total from the large pizzas?

Answer 1.2: He buys 2 large pizzas, so 2 * 16 = 32 slices. The answer is 32.

Question 1.3: How many slices are in one small pizza?

Answer 1.3: One small pizza has 8 slices. The answer is 8.

Question 1.4: How many slices are there in total from the small pizzas?

Answer 1.4: He buys 2 small pizzas, so 2 * 8 = 16 slices. The answer is 16.

Question 1.5: Now we can answer the question: How many pieces does he eat that day?

Answer 1.5: There are 32 slices from the large pizzas and 16 slices from the small pizzas, so he eats 32 + 16 = 48 pieces that day. The answer is 48.
}

Question 2: Josh decides to try flipping a house.  He buys a house for \$80,000 and then puts in \$50,000 in repairs.  This

increased the value of the house by 150\%.  How much profit did he make?

Question 2.1:
\end{prompt}
\begin{prompt}{A reward generation prompt for GSM8K}

Given a question and some sub-questions, determine whether the last sub-question is useful to answer the question. Output 'Yes' or 'No', and a reason.

{\promptEmphExample
Question 1: Four years ago, Kody was only half as old as Mohamed. If Mohamed is currently twice as 30 years old, how old is Kody?

Question 1.1: How old is Mohamed?

Question 1.2: How old was Mohamed four years ago?

New question 1.3: How old was Kody four years ago?

Is the new question useful? Yes. We need the answer to calculate how old is Kody now.
}

Question 2: Josh decides to try flipping a house.  He buys a house for \$80,000 and then puts in \$50,000 in repairs. This increased the value of the house by 150\%.  How much profit did he make?

New question 2.1: How much did Josh spend on the house?

Is the new question useful?

\end{prompt}
\vspace{-5mm}
\caption{
    Example prompts for GSM8K.
    {\promptEmphExample Italic texts} denote 1-shot examples.
}
\label{fig:appendix-prompt-gsm}
\end{figure*}

\begin{figure*}[hp!]
\begin{prompt}{An answer generation prompt for ARC}
Given a question, please decompose it into sub-questions. For each sub-question, please answer it in a complete sentence, ending with "The answer is". When the original question is answerable, please start the subquestion with "Now we can answer the question with an option from A to D: ".

{\promptEmphExample

Question 1: Juan and LaKeisha roll a few objects down a ramp. They want to see which object rolls the farthest. What should they do so they can repeat their investigation? Options: A) Put the objects in groups, B) Change the height of the ramp, C) Choose different objects to roll, D) Record the details of the investigation.

Question 1.1: What is necessary to ensure that experimental results can be repeated?

Answer 1.1: To ensure repeatability, experimental details must be accurately recorded. The answer is to record details.

Question 1.2: What kind of information should Juan and LaKeisha record for repeatability?

Answer 1.2: They should record details like the objects used, ramp height, and surface conditions. The answer is experimental conditions.

Question 1.3: How would recording experimental details help in the investigation?

Answer 1.3: Recording details allows them to recreate the exact same conditions for reliable comparison. The answer is that it enables consistent replication.

Question 1.4: Now we can answer the question with an option from A to D: What should they do to repeat their investigation?

Answer 1.4: Record the details of the investigation. The answer is D.
}

Question 2: Which method is the safest way to watch an eclipse of the Sun? Options: A)  Turn away after two or three minutes. B)  Look at the Sun through a long telescope. C)  Cast an image through a pinhole onto a screen. D)  Blink often until your eyes get used to the light..

Question 2.1: Why should you not look directly at the Sun during an eclipse?

Answer 2.1:

\end{prompt}

\begin{prompt}{An action generation prompt for ARC}

Given a question, please decompose it into sub-questions. For each sub-question, please answer it in a complete sentence, ending with "The answer is". When the original question is answerable, please start the subquestion with "Now we can answer the question with an option from A to D: ".

{
\promptEmphExample

Question 1: Juan and LaKeisha roll a few objects down a ramp. They want to see which object rolls the farthest. What should they do so they can repeat their investigation? Options: A) Put the objects in groups, B) Change the height of the ramp, C) Choose different objects to roll, D) Record the details of the investigation.

Question 1.1: What is necessary to ensure that experimental results can be repeated?

Answer 1.1: To ensure repeatability, experimental details must be accurately recorded. The answer is to record details.

Question 1.2: What kind of information should Juan and LaKeisha record for repeatability?

Answer 1.2: They should record details like the objects used, ramp height, and surface conditions. The answer is experimental conditions.

Question 1.3: How would recording experimental details help in the investigation?

Answer 1.3: Recording details allows them to recreate the exact same conditions for reliable comparison. The answer is that it enables consistent replication.

Question 1.4: Now we can answer the question with an option from A to D: What should they do to repeat their investigation?

Answer 1.4: Record the details of the investigation. The answer is D.

}

Question 2: Which method is the safest way to watch an eclipse of the Sun? Options: A)  Turn away after two or three minutes. B)  Look at the Sun through a long telescope. C)  Cast an image through a pinhole onto a screen. D)  Blink often until your eyes get used to the light..

Question 2.1:
\end{prompt}
\begin{prompt}{A reward generation prompt for ARC}

Given a question and some sub-questions, determine whether the last sub-question is useful to answer the question. Output 'Yes' or 'No', and a reason.

{
\promptEmphExample
Question 1: How are particles in a block of iron affected when the block is melted? Options: A) The particles gain mass, B) The particles contain less energy, C) The particles move more rapidly, D) The particles increase in volume.

Question 1.1: What happens to particle energy when a solid melts?

Question 1.2: How does the movement of particles change during melting?

New question 1.3: Why does increased movement signify a phase change to liquid?

Is the new question useful? Yes, because understanding why increased movement signifies a phase change helps clarify the behavior of particles during melting.
}

Question 2: Which method is the safest way to watch an eclipse of the Sun? Options: A)  Turn away after two or three minutes. B)  Look at the Sun through a long telescope. C)  Cast an image through a pinhole onto a screen. D)  Blink often until your eyes get used to the light..

New question 2.1: Why should you not look directly at the Sun during an eclipse?

Is the new question useful?

\end{prompt}
\vspace{-5mm}
\caption{
    Example prompts for ARC.
    {\promptEmphExample Italic texts} denote 1-shot examples.
}
\label{fig:appendix-prompt-arc}
\end{figure*}